\documentclass[sigconf,natbib=false, anonymous=False]{acmart}
\usepackage{enumitem}
\AtBeginDocument{%
  }

\setcopyright{acmcopyright}
\copyrightyear{2023}
\acmYear{2023}
\acmDOI{XXXXXXX.XXXXXXX}

\RequirePackage[
  datamodel=acmdatamodel,
  style=acmnumeric,
  ]{biblatex}

\usepackage{soul}
\begin{document}

\twocolumn[{%
\vspace{20mm}
{ \large
\begin{itemize}[leftmargin=2.5cm, align=parleft, labelsep=2cm, itemsep=4ex]

\item[\textbf{Citation}]{Z. Fowler, K. Kokilepersaud, M. Prabhushankar, G. AlRegib,  "Clinical Trial Active Learning," in \textit{Proceedings of the 14th ACM International Conference on Bioinformatics, Computational Biology and Health Informatics (ACM-BCB),} 2023.}

\item[\textbf{Review}]{Date of Acceptance: July 12th 2023}

\item[\textbf{Codes}]{\url{https://github.com/olivesgatech/Clinical-Trial-Active-Learning.git}}

\item[\textbf{Bib}]  {@inproceedings\{fowler2023clinical,\\
    title=\{Clinical Trial Active Learning\},\\
    author=\{Z. Fowler, K. Kokilepersaud, M. Prabhushankar, G. AlRegib\},\\
    booktitle=\{The 14th ACM Conference on Bioinformatics, Computational Biology and Health Informatics (ACM-BCB)\},\\
    year=\{2023\}\}}

\item[\textbf{Contact}]{
\{zfowler3, kpk6, mohit.p, alregib\}@gatech.edu\\\url{https://ghassanalregib.info/}\\}

\end{itemize} }}]
\newpage

%%
%% The "title" command has an optional parameter,
%% allowing the author to define a "short title" to be used in page headers.
\title{Clinical Trial Active Learning}

%%
%% The "author" command and its associated commands are used to define
%% the authors and their affiliations.
%% Of note is the shared affiliation of the first two authors, and the
%% "authornote" and "authornotemark" commands
%% used to denote shared contribution to the research.
\author{Zoe Fowler}
%\authornotemark[1]
\email{zfowler3@gatech.edu}
\affiliation{%
  \institution{Georgia Institute of Technology}
  \city{Atlanta}
  \state{Georgia}
  \country{USA}
}

\author{Kiran Kokilepersaud}
\email{kpk6@gatech.edu}
\affiliation{%
  \institution{Georgia Institute of Technology}
  \city{Atlanta}
  \state{Georgia}
  \country{USA}
}

\author{Mohit Prabhushankar}
\email{mohit.p@gatech.edu}
\affiliation{%
  \institution{Georgia Institute of Technology}
  \city{Atlanta}
  \state{Georgia}
  \country{USA}
}

\author{Ghassan AlRegib}
\email{alregib@gatech.edu}
\affiliation{%
  \institution{Georgia Institute of Technology}
  \city{Atlanta}
  \state{Georgia}
  \country{USA}
}

%%
%% By default, the full list of authors will be used in the page
%% headers. Often, this list is too long, and will overlap
%% other information printed in the page headers. This command allows
%% the author to define a more concise list
%% of authors' names for this purpose.
\renewcommand{\shortauthors}{Fowler et al.}

%%
%% The abstract is a short summary of the work to be presented in the
%% article.
\begin{abstract}
This paper presents a novel approach to active learning that takes into account the non-independent and identically distributed (non-i.i.d.) structure of a clinical trial setting. There exists two types of clinical trials: retrospective and prospective. Retrospective clinical trials analyze data after treatment has been performed; prospective clinical trials collect data as treatment is ongoing. Typically, active learning approaches assume the dataset is i.i.d. when selecting training samples; however, in the case of clinical trials, treatment results in a dependency between the data collected at the current and past visits. Thus, we propose prospective active learning to overcome the limitations present in traditional active learning methods and apply it to disease detection in optical coherence tomography (OCT) images, where we condition on the time an image was collected to enforce the i.i.d. assumption. We compare our proposed method to the traditional active learning paradigm, which we refer to as retrospective in nature. We demonstrate that prospective active learning outperforms retrospective active learning in two different types of test settings.
\end{abstract}

%%
%% The code bfrddddddddddddddddddddddddddddddddddddddddddddddddddddelow is generated by the tool at http://dl.acm.org/ccs.cfm.
%% Please copy and paste the code instead of the example below.
%%
\begin{CCSXML}
<ccs2012>
<concept>
<concept_id>10010147.10010257.10010282.10011304</concept_id>
<concept_desc>Computing methodologies~Active learning settings</concept_desc>
<concept_significance>500</concept_significance>
</concept>
<concept>
<concept_id>10010405.10010444.10010449</concept_id>
<concept_desc>Applied computing~Health informatics</concept_desc>
<concept_significance>300</concept_significance>
</concept>
</ccs2012>
\end{CCSXML}

\ccsdesc[500]{Computing methodologies~Active learning settings}
\ccsdesc[300]{Applied computing~Health informatics}

%%
%% Keywords. The author(s) should pick words that accurately describe
%% the work being presented. Separate the keywords with commas.
\keywords{Active learning, Deep learning, Retinal disease detection, Prospective clinical trials}

\received{11 June 2023}
\received[revised]{6 July 2023}
\received[accepted]{12 July 2023}

%%
%% This command processes the author and affiliation and title
%% information and builds the first part of the formatted document.
\maketitle

\section{Introduction}
\begin{figure}[h]
\centering
\includegraphics[width=0.45\textwidth]{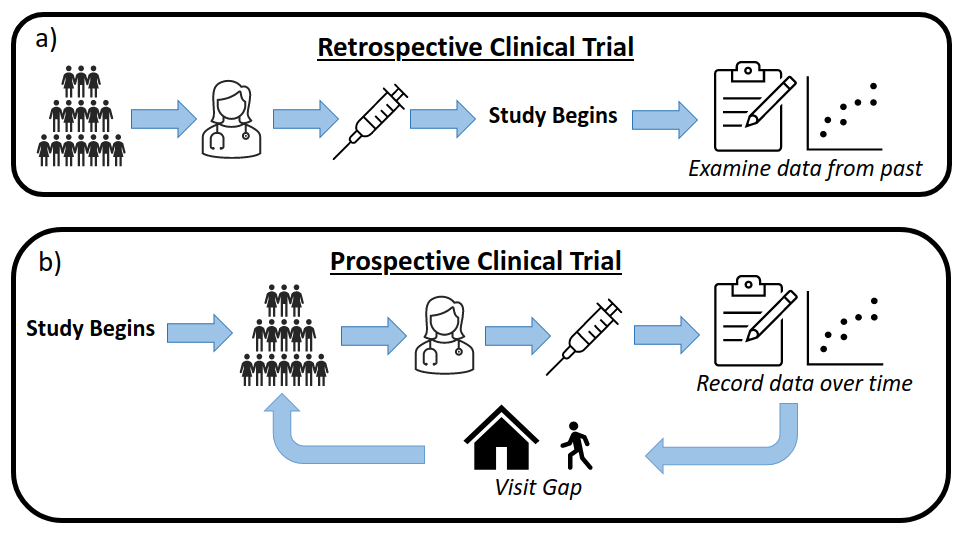}
\caption{Retrospective vs Prospective Clinical Trial Description}
\label{fig:retro_vs_prosp}
\end{figure}

Clinical trials are research studies that evaluate the effects of medical treatments \cite{chien2022multi}. There are two different types of clinical trials: retrospective and prospective as illustrated in Figure \ref{fig:retro_vs_prosp}. In part a) of Figure \ref{fig:retro_vs_prosp}, we show a retrospective clinical trial, where data is collected about participants after treatment has been administered. This type of clinical trial has access to the entirety of the pre-existing data and has no control over the design of the data collection process itself; it relies only on previous data to draw conclusions. However, part b) of Figure \ref{fig:retro_vs_prosp} shows that a prospective clinical trial collects data about participants over time as they are receiving treatment \cite{euser2009cohort}. Prospective clinical trials are designed with a specific treatment and research plan, and only the data collected through this specific clinical trial is used for analysis.

% In the case of ophthalmology, an example of a prospective clinical trial is illustrated in Figure \ref{fig:clinical}. In part a) of Figure \ref{fig:clinical}, the patient is assessed by a doctor at each visit. During this visit, the patient is examined, as shown in part b), and clinical values such as Best Corrected Visual Acuity (BCVA) and central subfield thickness (CST) are obtained. Then, diagnostic imaging, such as Optical Coherence Tomography (OCT), is collected from the patient. Using the collected information, the doctor recommends a treatment plan to manage Diabetic Retinopathy (DR) or Diabetic Macular Edema (DME), demonstrated in parts c) and d) of the figure. Finally, in part e) the patient follows up with the doctor after a few weeks to reassess the status of the disease. Throughout this process, data collected at each visit from a patient is dependent on factors from that patient's previous visits, mainly whether treatment was administered and how long the patient waits between successive visits.
\begin{figure*}[h]
    \centering
    \includegraphics[width=\textwidth]{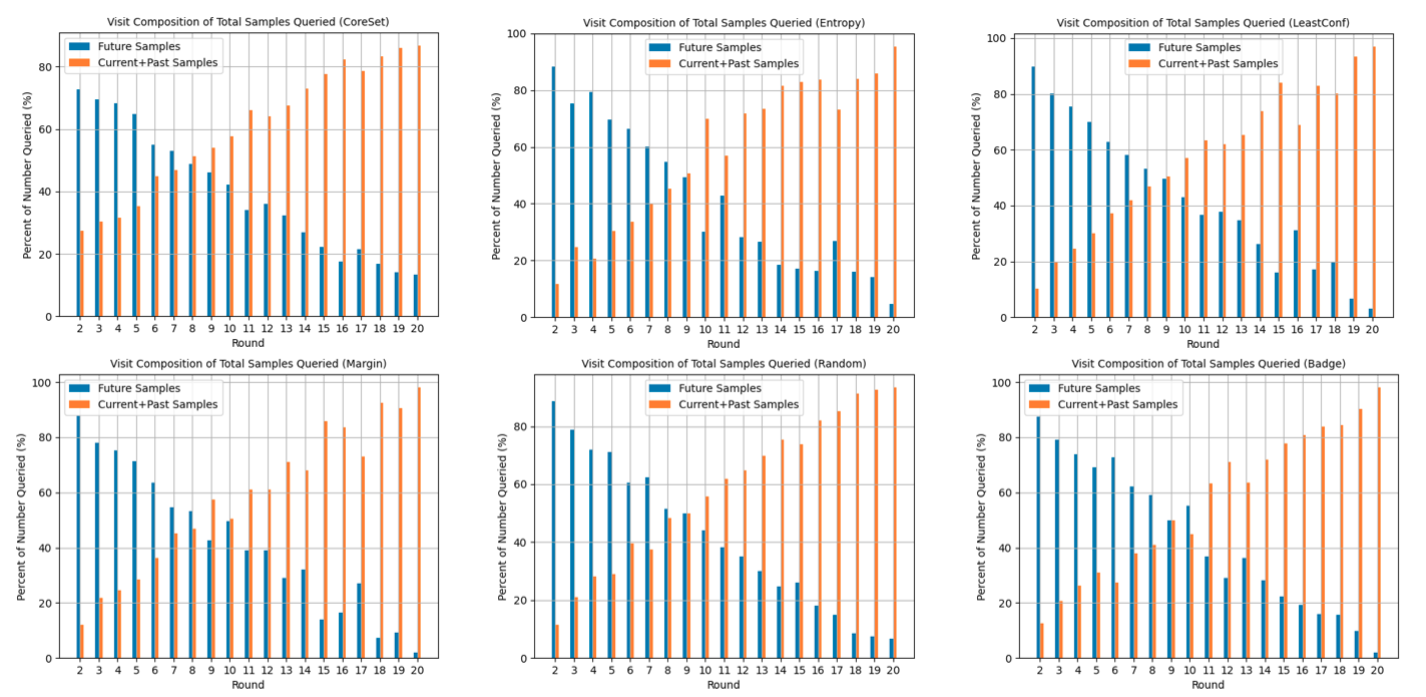}
    \caption{Retrospective active learning: For each query strategy, the percentage of samples queried at each training round from current and past visits versus samples chosen from future visits is compared. }
    \label{fig:unrealistic_sampling}
\end{figure*}

The prospective clinical trial procedure illustrated in Figure \ref{fig:retro_vs_prosp} has several complications. The main complication is that doctors assist a large volume of patients on any given day, leading to an increased demand when diagnosing patients. This demand is demonstrated in a study from 2015 \cite{mcdonald2015effects}, which concluded that the average radiologist is tasked with interpreting an image every 3-4 seconds to satisfy their large workload. This time requirement is typically not possible due to the time-consuming and difficult nature of interpreting these images. This increased workload leads to burnout, fatigue, and a large labeling error rate. 
% ML -> ML for medicine

To alleviate these complications, machine learning is a potential solution to assist doctors through many of the steps of a clinical trial. Machine learning identifies meaningful features from data automatically that can be used to obtain diagnostic decisions \cite{erickson2017machine}. Thus, machine learning allows model outputs to provide a second opinion to doctors when diagnosing patients for certain diseases \cite{kokilepersaud2022gradient}, \cite{kokilepersaud2023clinically}, \cite{wang2017chestx}, \cite{wu2019deep}. 

% In Figure \ref{fig:clinical}, the labeling task is eye disease classification, which affects treatment decisions and overall patient health. Given how crucial it is for medical data to be properly labeled and diagnosed, machine learning algorithms can be considered as a solution to automate many of the steps listed in Figure \ref{fig:clinical} of a clinical trial. Machine learning identifies meaningful features from data automatically, translating to completing labeling tasks at a persistent rate, unlike what is possible for human annotators. To support machine learning's potential, there has been a history of machine learning algorithms being applied successfully in the medical field, from detecting thoracic diseases in chest x-rays \cite{wang2017chestx} to screening for breast cancer \cite{wu2019deep}.
% problems between standard ML and medical ML
Applying standard machine learning approaches to the medical field is hardly straightforward. Specifically, there are two core challenges: 1) a large labeled dataset requirement and 2) an i.i.d. assumption of the dataset structure. Addressing the first point, most standard machine learning algorithms require large amounts of labeled data samples in order to accurately train a model. In the medical field, this labeled medical data can take many forms, a common one being disease diagnosis. However, as stated previously, this label is oftentimes time-consuming and challenging to provide due to the requirement of a trained medical expert.
% The second core problem relates to the structure of the medical data itself. Most machine learning algorithms assume that the dataset is independent and identically distributed (i.i.d.), disregarding information and an understanding of how data samples may be correlated across time \cite{scholkopf2021toward}. However, when analyzing part b) of Figure \ref{fig:retro_vs_prosp}, the data collected about a patient over time relies on the past data collected from that patient. In other words, data collected from prospective clinical trials is non-i.i.d., which is not accounted for in most standard machine learning approaches.
%% bar graph

Active learning is a common approach for tackling the first issue of needing large amounts of labeled data. Specifically, active learning does not require a large labeled set to train a machine learning model; instead, it is an iterative process that selects a smaller subset of training samples across multiple training rounds. Active learning uses query strategies to acquire this new selection of data and provides it to an external expert to label, reducing the overall amount of data annotated by an expert while still achieving better diagnosis with high accuracy \cite{settles2009active}. The query strategies aim to select the most informative samples, resulting in a refined training set that quickly reduces the generalization error \cite{kumar2020active}. In addition, active learning algorithms mirror the data collection process in prospective clinical trials as illustrated in Figure \ref{fig:retro_vs_prosp}. Namely, a patient deposits data on each visit, and an expert must choose a subset of this data to annotate based on a fixed labeling budget. This parallels the active learning paradigm, where the most informative samples are selected by a query strategy and labeled by an expert for each training round.
% One of the key similarities is that each patient visit can be thought of as a training round in the active learning paradigm, where this new patient data is added to the medical expert's or model's understanding of the studied diseases.
% One of the key similarities is that each patient visit can be equated to a training round in the active learning paradigm. This new \hl{This is incomplete}

% \begin{figure*}[h]
%     \centering
%     \includegraphics[width=\textwidth]{Figures/clinical_trial_new.png}
%     \caption{Clinical trial setup for DR/DME treatment}
%     \label{fig:clinical}
% \end{figure*}
% \begin{figure}[h]
%     \centering
%     \includegraphics[width=0.5\textwidth]{Figures/CHANGE_in_clinical.png}
%     \caption{Change in average BCVA and CST reported in patients across visits}
%     \label{fig:change_clinical}
% \end{figure}
% Specifically, prospective clinical trials test the effect of a treatment on a patient population over time; therefore, receiving treatment may result in a patient's disease status improving or worsening across visits. The time between visits can also result in a similar effect.
Despite active learning bypassing the need for large amounts of labeled data, prior query strategies assume that the data is i.i.d., as data is sampled based on solely a query criteria \cite{settles2009active}, \cite{wang2015querying}.  This leads to the second core problem: traditional active learning attempts make i.i.d. assumptions about the dataset structure, disregarding information about how data samples may be correlated across time \cite{scholkopf2021toward}. There are several issues with active learning failing to account for this problem: 1) active learning disregards the non-i.i.d. dependencies and consequences of these dependencies that arise during the clinical trial procedure and 2) active learning does not mirror how the data collection process is realistically performed in clinical trials. Examining the first point, part b) of Figure \ref{fig:retro_vs_prosp} illustrates that patients sometimes receive treatment at each visit during a prospective clinical trial. Patients then experience a gap between their next visit, where this process is repeated. Therefore, due to the treatment and the gap between visits, the manifestation of the disease changes between visits. For some patients, they may become more healthy, while others may become more ill. Hence, a patient's data at a particular visit can be viewed as a function of this previous visit information and cannot be treated as an independent data sample.
 \begin{figure*}[h]
\centering
\includegraphics[width=\textwidth, height=20.5em]{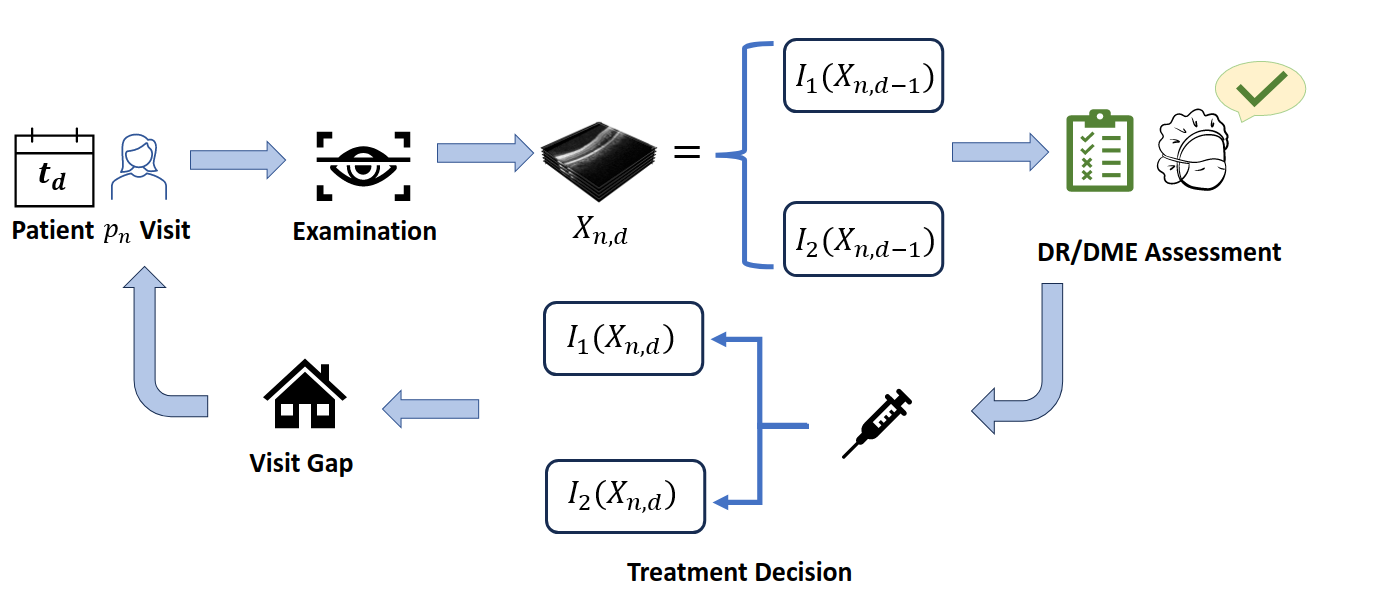}
\caption{Prospective clinical trial data collection process}
\label{fig:clinical}
\end{figure*}

% This is illustrated in Figure \ref{fig:change_clinical}, which demonstrates the shift in patients' BCVA and CST values across visits. Retrospective active learning does not account for this dependency; OCT images are selected regardless of which visit the images are from. 
%It makes sense then that in a real clinical setting, if the patients' disease manifestation shifts over time, it is expected that the deep learning model must also change to adapt to these different conditions.
% For datasets and real-life scenarios that contain data that is non-i.i.d., retrospective active learning can pose issues with regards to its sampling process. In the case of Figure \ref{fig:clinical} with OCT images, patients sometimes receive treatment at each visit. b/c it's non-iid, it does not mirror procedure\

The second challenge for using active learning in clinical trials is that the data collection process in clinical trials is itself non-i.i.d. At a specific visit in a clinical trial, the current patient information depends on the patient's medical data from previous visits. As a result, the sampling strategies deployed during traditional active learning algorithms are often unrealistic, as they sample images or data that the model would not necessarily be able to access at that point in a clinical scenario. This point is demonstrated in Figure \ref{fig:unrealistic_sampling}. In this figure, we train an active learning model on non-i.i.d. medical image data from the {\fontfamily{qcr}\selectfont OLIVES} dataset \cite{prabhushankar2022olives} for 20 rounds, adding in new queried data at each round. This figure shows the percentage of the total number of samples queried from future visits versus current and past visits.
% For example, at Round 1, we calculate the percentage of samples queried that are from any patient's Visit 1 data, as well as the percentage of samples queried that are from future visits. 
We repeat this calculation of the percentage at every training round for six different query strategies. As shown in Figure \ref{fig:unrealistic_sampling}, for all six query strategies the active learning model relies heavily on future samples until later rounds, meaning that active learning is \emph{not} being performed in a way that aligns with prospective clinical trials despite the routine itself matching the data collection process shown in Figure \ref{fig:retro_vs_prosp}. 

Traditional active learning algorithms further diverge from the prospective clinical trial process through their evaluation procedure. In a realistic prospective clinical trial setting, the treatment effect is evaluated on patients' current visit data. Thus, the data analyzed by medical experts is dynamic and changes at each visit. Traditional active learning algorithms, however, evaluate performance on a fixed test set across rounds, resulting in the evaluation procedure common in these traditional active learning paradigms to contribute further to how unrealistic these approaches are.
% here
Hence, traditional attempts to apply active learning to medical data can be considered as viewing the active learning process from a retrospective lens \cite{logan2022decal}, \cite{smailagic2018medal}. We argue instead that the \emph{active learning process for medical data must be viewed from the prospective clinical trial lens presented in Figure \ref{fig:retro_vs_prosp}} because of the nature of the clinical trial data collection process. Throughout this paper, we refer to traditional active learning methods as retrospective active learning, which assumes access to the entirety of the collected data with no concern as to how data samples may be correlated over time. On the other hand, we refer to prospective active learning as active learning methods that follow the structure of prospective clinical trials by querying data sequentially. To the best of our knowledge, this is the first work that uses machine learning in a prospective clinical trial setting.

In this paper, we address the aforementioned problems by introducing a prospective active learning methodology that overcomes the inherent limitations of existing active learning strategies to apply them in clinical settings. Specifically, we consider the data collection process of clinical trials that creates time-related dependencies across data samples in clinical trial datasets, and we examine different model test settings by considering realistic clinical setting conditions. Thus, the contributions of this paper are as follows:
\begin{enumerate}
  \item We convert both retrospective and prospective clinical trials into active learning frameworks by following the setup for both described in Figure \ref{fig:retro_vs_prosp}.
  \item We demonstrate and compare performance for this setup against retrospective active learning results on the classification of diseased Optical Coherence Tomography (OCT) scans from an ophthalmology prospective clinical trial.
  \item We analyze our prospective active learning framework in different test settings, including a fixed test set to demonstrate overall generalization capabilities and a dynamic test set that generalizes to specific visits.
\end{enumerate}

\section{Non-i.i.d. Explanation}

%%%%REWRITING
In this section, we discuss in depth the non-i.i.d. structure of clinical trial data. Consider an ophthalmology prospective clinical trial for Diabetic Retinopathy (DR) and Diabetic Macular Edema (DME) assessment, illustrated in Figure \ref{fig:clinical}. In this clinical trial, assume we have a set of patients $P = \{ p_{1}, \cdots , p_{N} \}$, where each $p_{n} \in P$ represents a unique patient. Additionally, each $p_{n}$ is associated with a set of visits $T = \{ t_{1}, \cdots , t_{D} \}$, where each $t_{d} \in T$.
Figure \ref{fig:clinical} indicates that at a particular visit $t_{d}$, a patient $p_{n}$ is assessed for DR and DME through a medical expert's examination of the patient's OCT data of the form $X_{n,d}$, which indicates the patient number $n$ and time step number $d$ the data originated from. Then, depending on the expert's assessment, patients receive an eye injection $I_{1}$ or $I_{2}$ depending on which disease the patient is diagnosed with. Thus, a patient's OCT scans at a particular visit are dependent on all previously administered treatment injections. We represent this as $X_{n,d} = I_{1}(X_{n,d-1})$ or $X_{n,d} = I_{2}(X_{n,d-1})$, as $X_{n,d}$ is a result of the treatment decision at patient $p_{n}$'s previous visit $t_{d-1}$. Because a patient's current visit data depends on the previous visit, the data collected at each visit stage cannot be assumed to be i.i.d. 

The i.i.d. assumption prevalent in machine learning algorithms expects that data samples from a particular dataset are independently drawn from the same probability distribution \cite{wu2022towards}. Throughout this paper, we consider a clinical trial dataset to be i.i.d. if a patient's data deposited at a particular visit has no dependency on previous visit data. However, in prospective clinical trial settings, the effect of a certain treatment is observed by analyzing a patient's data across several visits. By nature, the resulting clinical trial dataset is non-i.i.d., which retrospective active learning algorithms inherently disregard.

\begin{figure}[h]
\centering
\includegraphics[width=0.45\textwidth]{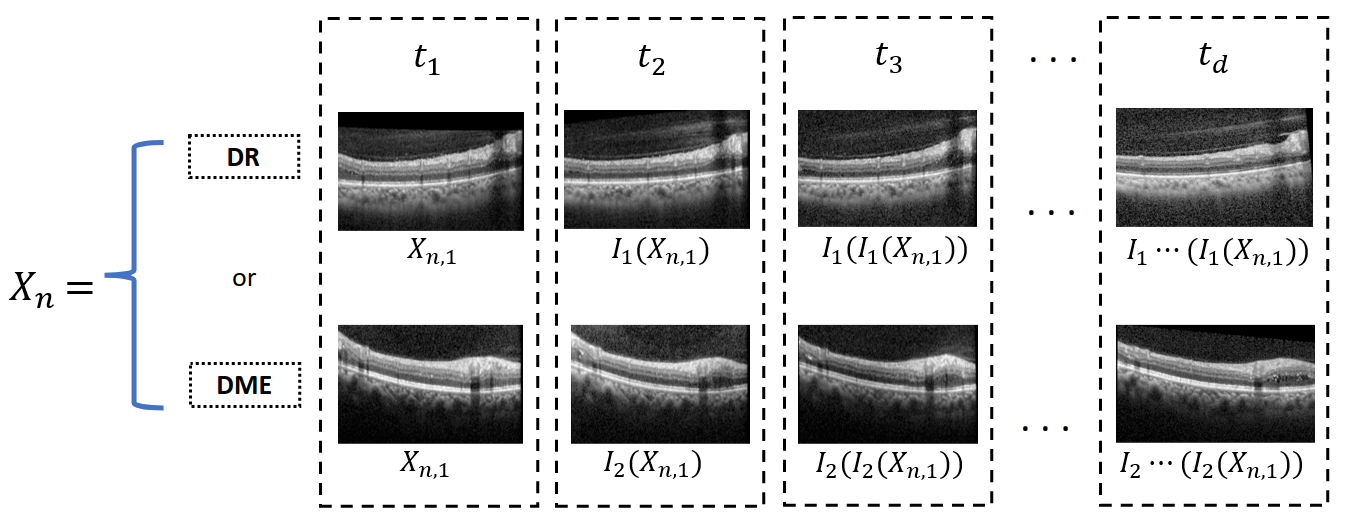}
\caption{Patient $p_{n}$'s total collected data $X_{n}$ across $d$ visits (Intervention function applied between successive visits is dependent on which disease class patient $p_{n}$ belongs to)}
\label{fig:xn}
\end{figure}

To understand how taking into account this non-i.i.d. structure affects active learning performance, we analyze the aforementioned data collection process through Figure \ref{fig:xn}. 
% Figure \ref{fig:xn} visualizes the total collected OCT images $X_{n}$ corresponding to all $d$ visits of a patient $p_{n}$ that has been diagnosed with either DR or DME. 
In Figure \ref{fig:xn}, assume a patient $p_{n}$ that has been diagnosed with either DR or DME has completed the clinical trial, resulting in a total of $d$ visits. Figure \ref{fig:xn} visualizes the total collected OCT images $X_{n}$ from this patient across their $d$ visits.
Between each time step or visit, a treatment, unique to the disease, has been administered. We term this an intervention and represent it using $I_{1}$ for DR or $I_{2}$ for DME. This intervention relates the OCT images at $t_{d}$ and $t_{d-1}$ for any $d \in \{1, \cdots, D\}$. In the context of classification, estimating the intervention function by utilizing machine learning algorithms is a viable way to classify a data sample as belonging to a particular class. 

% To estimate this intervention function, one might take the inverse of the random variables present between two successive time steps, or machine learning might be utilized to provide an estimate for this intervention function.

% by drawing an analogy with random variables. Assume the data collection process in Figure \ref{fig:clinical} produces a $N \times D$ dataset $X$, resulting from a total of $N$ patients with $D$ total visits. Taking a row from the dataset, we obtain $X_{n} = \{x_{n,1}, . . . , x_{n,D}\}$, where this row can be viewed as a realization of $D$ random variables. Furthermore, since the $N$ patients of dataset $X$ belong to one of two classes, $X$ contains realizations of two different random variables. Between each time step or visit $t$, an intervention function $I_{1}$ or $I_{2}$ unique to each class relates the random variables at $t_{d}$ and $t_{d-1}$ for any $d \in D$. In the context of classification, estimating the intervention function is a viable way to classify a data sample as belonging to a particular class. To estimate the intervention function, one might take the inverse of the random variables present between two successive time steps, or machine learning might be utilized to provide an estimate for this intervention function.

However, the overall order in which the data is analyzed to determine the intervention function affects the overall estimate of this function and thus classification decisions. We argue that if the data is observed in a sequential manner, the intervention function $I_{1}$ or $I_{2}$ will be able to be properly estimated. However, if the data is observed non-sequentially, the intervention estimated is likely not equivalent to the actual intervention, resulting in an incorrect classification result since the data at a particular visit is dependent on all previous intervention functions applied during past visits. By creating a framework that queries images prospectively, i.e. in a sequential manner based on patient visit number, we separate our work from previous active learning works that make an i.i.d. assumption on the data. 
We show that with this framework, we attain increased performance on a non-i.i.d. clinical trial dataset by exploiting the relationship between data at each time step.

% To illustrate this point, we refer to Figure \ref{fig:estimation-oned} which presents a one-dimensional, non-probabilistic viewpoint of this claim. In Figure \ref{fig:estimation-oned}, we show that by querying samples in a sequential fashion to add to our training set, we obtain very different results than by querying non-sequentially, and we better estimate the true intervention function.
% \begin{figure}[h]
% \centering
% \includegraphics[width=0.47\textwidth]{Figures/estim-oned.png}
% \caption{Explanation of the effect of querying sequentially}
% \label{fig:estimation-oned}
% \end{figure}

% Figure \ref{fig:estimation-oned} demonstrates that when querying data samples sequentially, the intervention function estimate is $x_{0}t_{d}$, the initial data sample value multiplied by the time step number. However, when data samples are non-sequentially added to our training set, a potential intervention function estimate for this set of data is $3x_{d-1}$, or 3 times the data sample value at the previous time step.

% and thus can be compared to the visualization in Figure \ref{fig:estimation-rv}. By creating a framework that queries images prospectively, i.e. in a sequential manner based on patient visit number, we separate our work from previous active learning works that make an i.i.d. assumption on the data. 
% We show that with this framework, we attain increased performance on a non-i.i.d. medical dataset by exploiting the relationship between data at each time step.

\section{Relation to Prior Work}

\subsection{Deep Learning in OCT}
The proposed framework is applicable to any clinical trial. However, we utilize OCT data as a case study. The clinical trial process can be laborious, as demonstrated in an ophthalmology setting through Figure \ref{fig:clinical}. Hence, there is an interest in applying deep learning to ophthalmology to reduce diagnosis time and increase diagnostic accuracy. For example, \cite{temel2019relative} uses a transfer learning-based approach to detect abnormalities in relative afferent pupillary screening. \cite{choudhary2023deep} is aimed towards using machine learning for accurate disease classification of four different retinal disease categories. \cite{dai2021deep} uses fundus images for labeling the severity of DR and the presence of other retinal disease factors. \cite{tsuji2020classification} uses a capsule network to exploit positional information in OCT images for disease classification. 

Other approaches have been designed to make deep learning more explainable from a medical viewpoint. For example, \cite{logan2022multi} uses multi-modal learning to incorporate OCT scans alongside physician insights for disease classification. Additional methods aim to uncover meaningful clinical insights and relationships. \cite{kokilepersaud2023clinically}, for instance, uses contrastive learning to discover a relationship between clinical values and biomarker structures that manifest within OCT scans. All the aforementioned methods demonstrate the vast potential for deep learning to be applied in ophthalmology settings, despite the requirement of a large set of labeled training data.

\subsection{ Medical Active Learning}
\begin{figure*}[h]
    \centering
    \includegraphics[width=\textwidth]{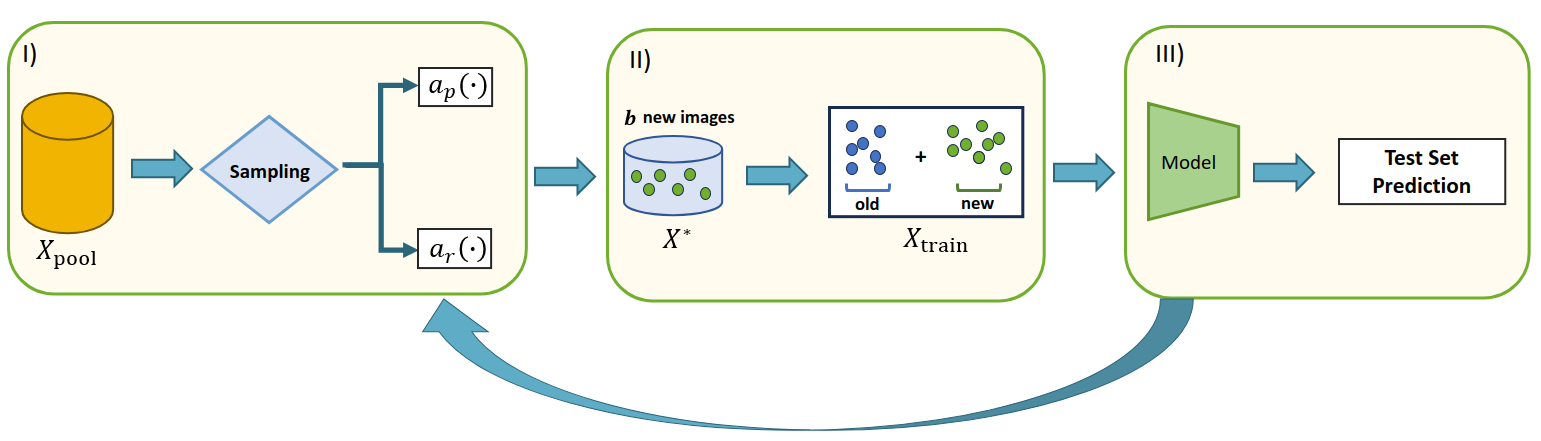}
    \caption{Active learning framework - I) Choose acquisition strategy and sample from $X_{\text{pool}}$ accordingly II) Add sampled data $X^{*}$ to the training set $X_{\text{train}}$ III) Evaluated trained model on test set}
    \label{fig:AL}
\end{figure*}
The premise behind active learning is that a machine learning algorithm can achieve higher accuracy with fewer labeled samples \cite{settles2009active}. By choosing the most relevant data samples, active learning is able to reduce the overall labeled data needed to train a model, while still experiencing increased model performance. Active learning is performed iteratively; in each iteration of training a machine learning model, a query strategy is applied to determine the most informative samples to label. These samples are then added to the model's training set. The simplest strategy is to select unlabeled data samples randomly, known as random sampling. Uncertainty-based methods are also popular and include entropy uncertainty, margin sampling, and least confidence sampling. Entropy presents a simple solution to querying samples by identifying the top $B$ most uncertain samples to label \cite{wang2014new}. Margin sampling chooses samples that have the smallest margin between the two highest predicted classes, and least confidence selects samples whose predicted labels have the lowest probability \cite{wang2014new}. Other methods for querying samples focus on enhancing sample diversity. These diversity-based strategies include Coreset, where new samples are queried at each round using furthest-first traversal on prior labeled samples \cite{sener2017active}, \cite{ash2019deep}. The BADGE query strategy \cite{ash2019deep}, on the other hand, takes into account both the diversity and uncertainty of samples. Active learning is favorable in situations where acquiring labeled data samples is challenging and time-consuming.

Using active learning in a medical setting is well-motivated by the fact that labeling medical images is expensive due to requiring a trained medical expert \cite{bangert2021active}. Therefore, it is essential to develop algorithms that reduce the labeling burden in order to assist medical experts in their practice. Thus, there have been many previous efforts to apply active learning for this purpose. Mainly, efforts revolve around altering active learning frameworks to drastically reduce the number of labeled images needed in order to achieve high accuracy. For example, \cite{bangert2021active} designed an active learning framework by adding a pre-labeling model to provide human annotators with estimated class labels to reduce overall labeling time. The authors applied this method in many medical case studies, one of them being the detection of Covid-19 in a dataset, and found that with their active learning algorithm, they needed only $5\%$ of the images to be human-labeled in order to achieve $93\%$ accuracy on the Covid-19 dataset. 

Other efforts to incorporate active learning in a medical setting and reduce the overall number of labeled images include \cite{smailagic2018medal}, \cite{logan2022decal}, \cite{logan2022patient}, \cite{el2022multi}, and \cite{hoi2006batch}. \cite{smailagic2018medal} aims to sample unlabeled images by combining entropy sampling and a distance function that maximizes the average distance to all remaining images in the data pool in the feature space. \cite{logan2022decal}, \cite{logan2022patient} provides a plug-in approach that combines patients' clinical data with existing active learning query strategies for the sampling of medical images. \cite{el2022multi} uses active learning to diagnose heart disease and tests generalization capability, and \cite{hoi2006batch} is an earlier work on active learning that uses the Fisher information matrix to query batches of unlabeled samples and tests this method's ability to categorize medical images.

The aforementioned methods have many complications in a real-life clinical setting. Mainly, all are evaluated retrospectively, with no discussion on whether the medical data the algorithms have been both trained and evaluated on exhibit any dependencies. All attempts consider each data sample as being independent from others, even if the dataset utilized is from prospective clinical trials.

\section{Proposed Method}
\subsection{Dataset}
\begin{figure*}[h]
    \centering
    \includegraphics[width=\textwidth]{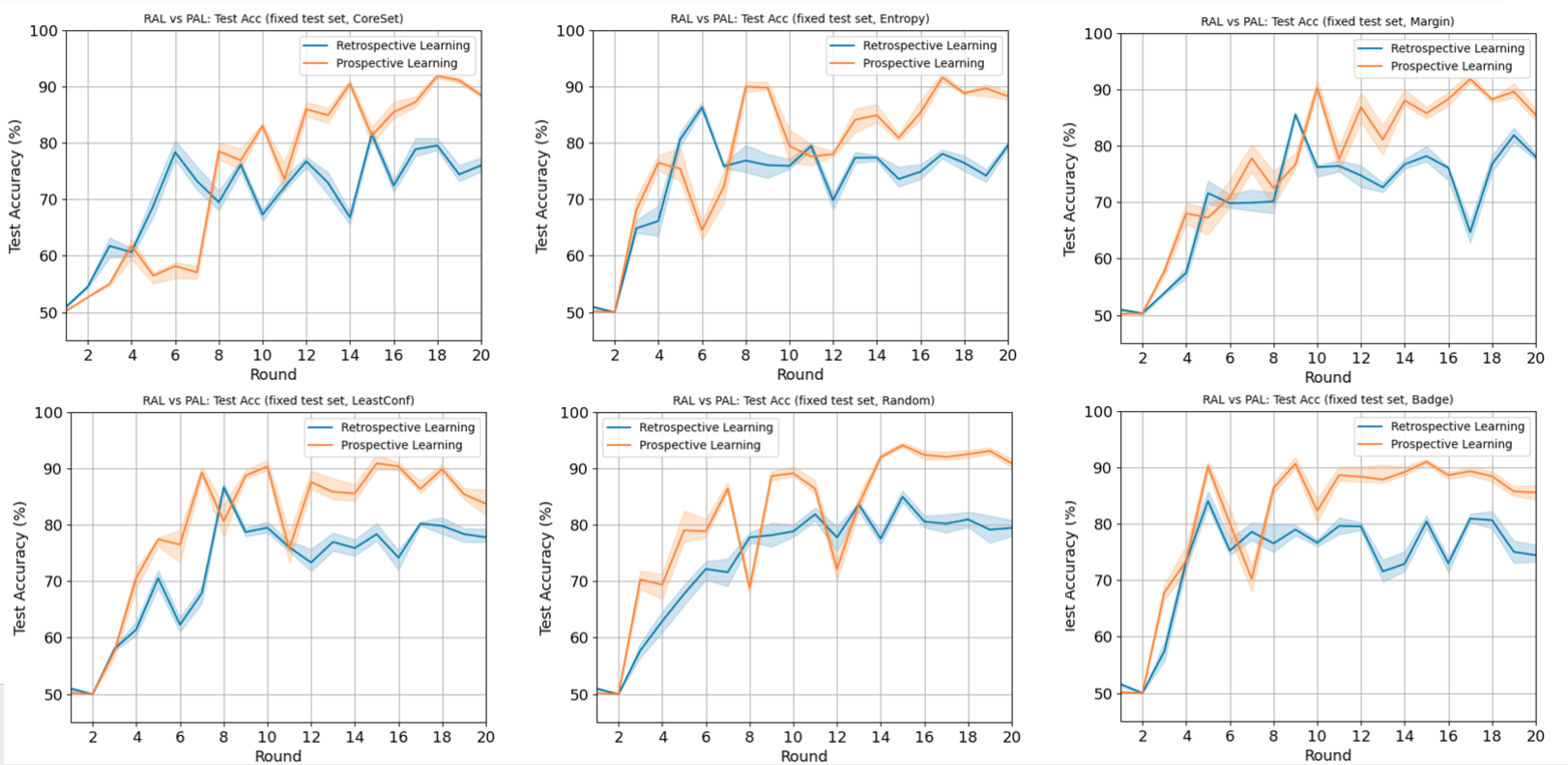}
    \caption{Retrospective active learning (RAL) vs Prospective active learning (PAL) test accuracy results averaged over 5 seeds. }
    \label{fig:test-acc}
\end{figure*}
For our experiments, we utilize the {\fontfamily{qcr}\selectfont OLIVES} \cite{prabhushankar2022olives} dataset. The {\fontfamily{qcr}\selectfont OLIVES} dataset consists of $78,189$ OCT images, collected from two different clinical trials. The two clinical trials are denoted as PRIME, which recruited patients with DR, and TREX, which recruited patients with DME. The patients in PRIME and TREX received two distinct treatments, allowing the assumptions from Figure \ref{fig:xn} to reflect the structure of the actual dataset. The DR patients in PRIME receive treatment $I_{1}$, whereas the DME patients in TREX receive a different treatment $I_{2}$.

The clinical trial format of the {\fontfamily{qcr}\selectfont OLIVES} dataset overall follows closely to what is described in Figure \ref{fig:clinical}. Patients arrive at a clinic and have a series of diagnostic exams and imaging procedures performed on them. Afterwards, they are evaluated for treatment and wait until their next follow-up appointment to repeat the data collection process. On average, patients in the dataset were enrolled in the clinical trials for approximately 114 weeks. On average, this equates to 16.6 visits per patient and 14.2 treatment injections. The number of weeks between visits varies across the clinical trials; in most cases, there are around 4 weeks between visits, though this amount increases if patients miss appointments or if the clinical trial protocol adjusts this length of time.

The {\fontfamily{qcr}\selectfont OLIVES} dataset reflects the clinical trial process by mapping each OCT image to a visit number that indicates when in time an image was collected during the clinical trials. Each OCT image also corresponds to a disease state (DR/DME), anonymous patient identity number, and anonymous eye identity number. The {\fontfamily{qcr}\selectfont OLIVES} dataset provides data from 96 unique eyes and 87 unique patients. From the 96 unique eyes, we derive our train and test sets. For the train set, we include 76 unique eyes and utilize the remaining 20 eyes for our test set. The test set is balanced between disease classes.

\subsection{Problem Formulation}
Assume access to a dataset $X$ that contains a time-series component such that the dataset can be represented as
\begin{equation}
    X = \{ x_{t_{1}}, x_{t_{2}}, \dots , x_{t_{D}} \},
\end{equation}
where $x_{t}$ represents the data collected at a specific time $t_{d}$ for any $d \in \{1, \cdots, D\}$ from all $N$ associated patients $P = \{ p_{1}, \cdots , p_{N} \}$. 
% Each patient $p_{n}$ is associated with a set of visits $T = \{ t_{1}, \cdots , t_{D} \}$.
% For each time step or visit number, we have data deposited during this specific visit from $N$ different patients: 
Thus, the resultant dataset has the form:
\begin{equation}
    \begin{aligned}
    x_{t_{1}} = \{ p_{1,1}, p_{1,2}, \dots , p_{1,N} \}\\
    \vdots\\
    x_{t_{D}} = \{ p_{D,1}, p_{D,2}, \dots , p_{D,N} \}
    \end{aligned}
\end{equation}

Consider the context of active learning, where the objective is to iterate through the unlabelled data pool $X_{\text{pool}}$ of dataset $X$ and select a subset of data to add to the training set $X_{\text{train}}$. At each round of training, query strategies select a batch of $b$ data samples $X^{*}$ until the budget constraint $B$ is met. Following our definition of retrospective active learning, we can adapt \cite{benkert2023gaussian} to define the selection of such samples in the retrospective case as
\begin{equation}
    X^{*} = \underset{x_{t_{1}}, \dots , x_{t_{b}} \in X_{\text{pool}}}{\arg\max} a_{retro}(x_{t_{1}}, \dots , x_{t_{b}})
    % \arg\max_{x_{1}, \dots , x_{b} \in X_{\text{pool}}} a(x_{1}, \dots , x_{b})
\end{equation}
where $a_{retro}$ is the acquisition function for the retrospective setting. In the case of retrospective active learning, this is simply a pre-existing query strategy (i.e., random, least confidence, etc.) that selects data from $X_{\text{pool}}$ based off of a specific criteria unique to each query strategy.

For prospective active learning, the non-i.i.d. structure of the data is taken into consideration. Our selection of samples in this case is
\begin{equation}
    X^{*} = \underset{x_{t_{1}}, \dots , x_{t_{b}} \in X_{\text{pool}}}{\arg\max} a_{pro}(x_{t_{1}}, \dots , x_{t_{b}}  | d)
    % \arg\max_{x_{1}, \dots , x_{b} \in X_{\text{pool}}} a(x_{1}, \dots , x_{b} | T)
\end{equation}
where we augment Equation 4 such that we condition the selection based on the time or visit number $d$ of when the data was collected.  This results in $a_{pro}$, which constrains the selection space on which pre-existing strategies can operate.

We design and implement an active learning framework to test the effect of querying images in a prospective manner using our formulation of $a_{retro}$ and $a_{pro}$. We apply this framework to disease classification using the {\fontfamily{qcr}\selectfont OLIVES} dataset, where OCT images from the dataset are classified as having either DR or DME. The OLIVES dataset, like clinical trials, provides visit-wise data. Hence, at the first visit, OCT scans will be collected from $N$ patients, resulting in $N\times 49$ OCT scans. Therefore, when performing prospective active learning via $a_{pro}$, we only have $N\times 49$ OCT scans to sample from in the first round. For the next visit, there are the remaining OCT scans from the first visit that were not selected during the first round, as well as an additional $N\times 49$ OCT scans corresponding to the patients’ second visit, that can be sampled from during the second round of training the prospective active learning model. In other words, each round $d$ of training an active learning model corresponds to patients’ visit $d$ and past (unused) visit data. On the other hand, retrospective active learning will have access to the $N\times 49$ OCT scans from the first visit, the $N\times 49$ OCT scans from the second visit, and so forth. Hence at the first round itself there are $D\times N\times 49$ OCT scans to choose from, where $D$ represents the total number of visits recorded in the dataset. At the second round of training the model, retrospective active learning has access to all the unused OCT scans from the previous round. The retrospective case presents all OCT images in the training pool as an option to be queried from, regardless of which patient visit they were collected at.

We use Figure \ref{fig:AL} to describe this experimental setup. Following the earlier discussion, in part I of Figure \ref{fig:AL} we iterate through $X_{\text{pool}}$ of the dataset, sampling our b data samples $X^{*}$ prospectively via $a_{pro}$ or retrospectively via $a_{retro}$. Once this new data has been queried, we add this data to the training set $X_{\text{train}}$ in part II of Figure \ref{fig:AL}. Afterwards, part III of Figure \ref{fig:AL} shows the trained model being evaluated on the test set for DR and DME classification. Finally, we repeat these steps, where we query new images either prospectively or retrospectively at each round of training until a total of $B$ images have been queried, with $B$ describing the overall budget constraint.

\begin{figure*}[h]
    \centering
    \includegraphics[width=\textwidth]{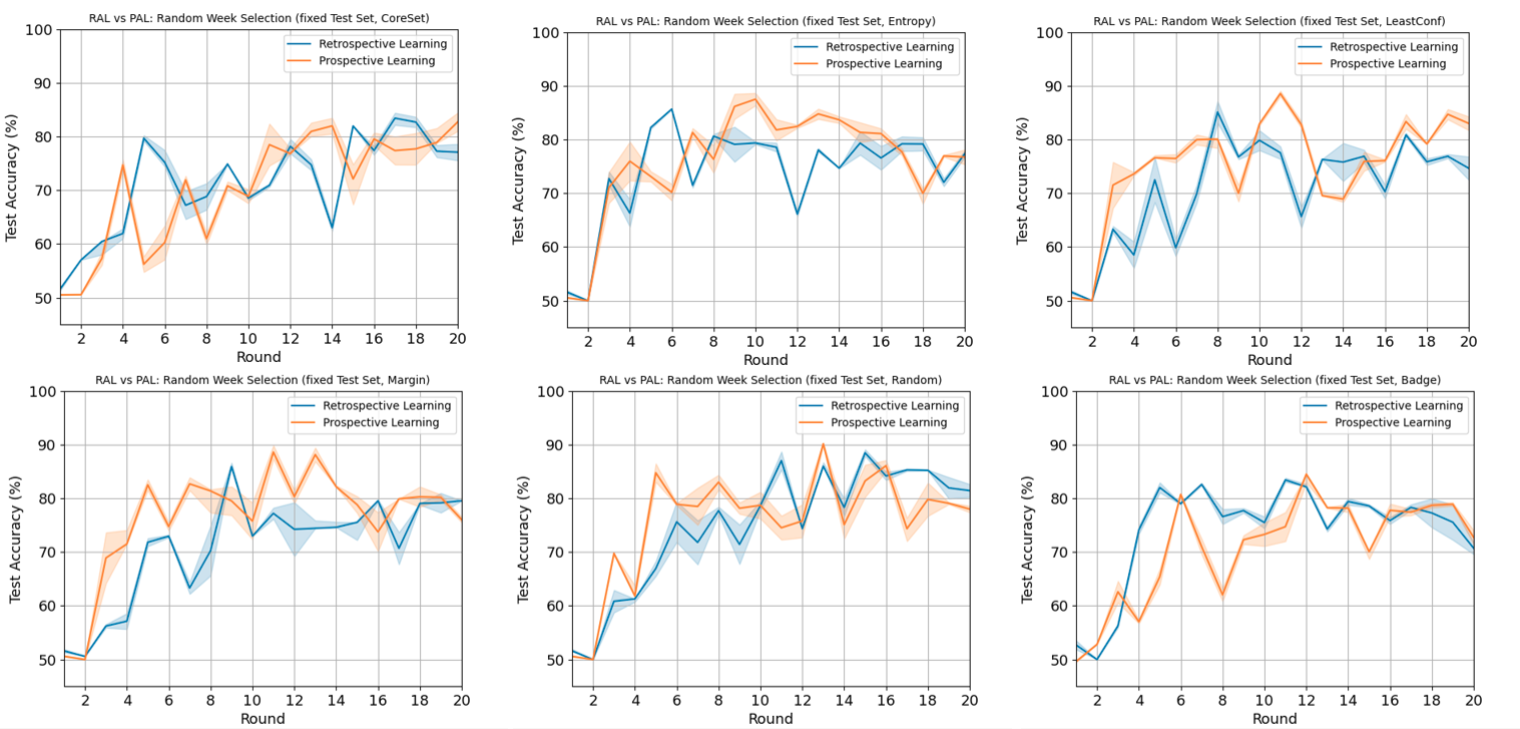}
    \caption{Retrospective active learning (RAL) vs Prospective active learning (PAL) test accuracy results averaged over 5 seeds, where PAL queries visits non-sequentially. }
    \label{fig:test-acc-rand}
\end{figure*}

\section{Experimental Design and Results}
\subsection{Experimental Design}
We implement the active learning framework described in Figure \ref{fig:AL}. We initialize the model's training set with 128 randomly selected data samples. At each successive training round $d$, 256 new data samples are queried. These data samples are selected either retrospectively or prospectively. The model's training set then consists of these newly queried samples in addition to samples that were selected in previous rounds. After the model has been trained with these new samples, the model is evaluated on the test set. This process is then repeated for $20$ rounds (corresponding to a total of 20 visits), resulting in a budget $B = 4,992$ images. With the current query size selection of 256 images, the medical expert needs to label only 6.38$\%$ of the total data collected, as the OLIVES dataset contains a total of 78,189 OCT images. Our current query size selection takes into account a medical expert’s limited labeling budget while still providing enough training data at each round such that the model is able to learn the fine-grained differences between DR and DME.

For our experiments, we utilize the Pytorch implementation of the Resnet18 architecture \cite{he2016deep}, with the linear layer being altered to accommodate binary classification. Thus, the input into the Resnet18 model is a batch of OCT images, and the output is a binary classification decision corresponding to the presence of either DR or DME. We train this model each round for DR/DME detection until a minimum accuracy of 97$\%$ is met. The Adam optimizer was used during training, with a learning rate of 0.00015. This training regimen was adopted from \cite{prabhushankar2022olives} and \cite{logan2022patient}, which presented results for DR/DME classification using these hyperparameter values. All OCT images were resized to $128\times128$ and were normalized with $\mu = 0.1706$ and $\sigma = 0.2112$.

We define two different test sets to examine the impact of querying images prospectively: 1) a fixed test set and 2) a dynamic test set. The fixed test set aims to test the trained model's overall generalization capabilities, whereas the dynamic test set examines the trained model's current and past visit generalization performance. We differentiate between the two test sets as follows: 
\subsubsection{Fixed Test Set}
The fixed test set is composed of $2,000$ OCT images from the 20 eyes excluded from the train set. The fixed test set is balanced between the two disease classes and contains OCT images collected from patients' Visit 1 through Visit 22 data. After each round of training both the prospective and retrospective active learning model, the trained model is evaluated on this particular test set. This test set evaluates how well the model can distinguish between two different eye disease classes across time and varying levels of severity.

\subsubsection{Dynamic Test Set}
The dynamic test set increases in size after each round of training the active learning model. At each round $d$ of training, 80 OCT images collected at visit $d$ from the 20 eyes excluded from the train set are added to the dynamic test set. At the final round of training, this results in a dynamic test set size of a total of 1,600 OCT images. The OCT images added at each round are class-balanced, and the composition of the dynamic test set is the same when evaluating the performance of the retrospective and prospective approaches. 

In a prospective clinical trial setting, the treatment effect is evaluated only on patients' current visit data. Thus, the total amount of data that has been evaluated by medical experts increases at each patient visit. The dynamic test set contributes to realistically assessing the performance of both retrospective and prospective active learning in a clinical trial setting, as the model at each round is evaluated on data from that current visit in addition to previous visits. The dynamic test set ensures that both active learning setups are able to generalize to the current visit, while not losing important information collected about previous visits. 

% here!
\subsection{Fixed Test Set Results}
\subsubsection{Retrospective vs Prospective Active Learning}
A comparison between retrospective and prospective active learning for the fixed test set is shown in Figure \ref{fig:test-acc} for the following query strategies: Random, Entropy, Margin, Least Confidence, Coreset, and BADGE. Results are averaged over 5 seeds and presented in Figure \ref{fig:test-acc}. In these plots, the x-axis corresponds to the number of training samples queried at each round, while the y-axis corresponds to the performance accuracy on the test set. The standard errors for each method are shown by the shaded regions.

For majority of the query strategies, the prospective active learning setting outperforms the traditional retrospective setting, particularly during later rounds. Because prospective active learning queries data in a sequential manner, we hypothesize that prospective active learning distinguishes between the two types of treatment administered to the two disease classes, resulting in better classification decisions. To further validate this hypothesis, we assess the impact of the order in which visits are integrated into the sampling process in prospective active learning.
% From the test accuracy results presented in Figure \ref{fig:test-acc}, it appears that once the model acquires around 2,400 samples, corresponding to patients' ninth visit, the prospective active learning method consistently outperforms the retrospective method for most query strategies, suggesting that the first nine visits may be crucial to model generalization and performance. 
We define a new experimental setup, where the prospective active learning framework randomly queries new visits non-sequentially at each round. Results for this setup are presented in Figure \ref{fig:test-acc-rand}.

As demonstrated in Figure \ref{fig:test-acc-rand}, prospective active learning fails to consistently outperform traditional retrospective active learning in this experimental setup, implying that the order in which these images are added to the model's training pool is significant. In particular, it implies that the model must receive the images consecutively in order to provide accuracy benefits over retrospective active learning. This indicates that each patient visit is dependent on the last visits and can not be treated independently, as is what is assumed in typical active learning experiments. Thus, by adopting a prospective active learning framework in this medical setting, the model is able to understand and learn the interventions on the training data at each visit, resulting in higher performance and generalization ability. 

When analyzing Figure \ref{fig:test-acc}, we also notice that at Round 8, there is a clear transition in efficiency between prospective and retrospective active learning across all query strategies. The transition in efficiency between the two methods is most likely due to the increased training set size at that particular round. In Figure \ref{fig:test-acc}, Round 8 corresponds to 1,920 training images, as the model is initialized with 128 images and queries 256 new images at successive rounds. In previous rounds, it is harder to generalize well to the test set due to a smaller training set size. This is because medical data has fine-grained structures that are difficult for a model to understand \cite{logan2022decal}, \cite{kokilepersaud2023clinically}; therefore, a larger initial training set is needed for the model to accurately classify an image. In addition, prospective active learning converges much quicker than retrospective active learning. Figure \ref{fig:test-acc} shows that the model does not necessarily need to train for the full 20 rounds in order to perform well. The BADGE and Margin query strategies, for instance, indicate that with prospective active learning, achieving a test accuracy of over $90\%$ can be achieved as early as rounds 9 or 10.

In Figure \ref{fig:test-acc}, we note that prospective active learning less consistently outperforms retrospective active learning for some query strategies, such as Random and Coreset. Potential reasons for the less consistent regions in performance accuracy come from the nature of the query strategies themselves. For example, Coreset relies on a learned representation of the data to query from; thus, performance in the beginning can be low because it may be utilizing poor representations of data to decide which samples to query \cite{muthakana2019uncertainty}. Figure \ref{fig:unrealistic_sampling} demonstrates that in retrospective active learning, Coreset is mainly querying from future data samples, thus allowing the model to access more diverse data, which results in an initial performance increase over prospective active learning. 

\begin{figure}[h]
\centering
\includegraphics[width=0.45\textwidth]{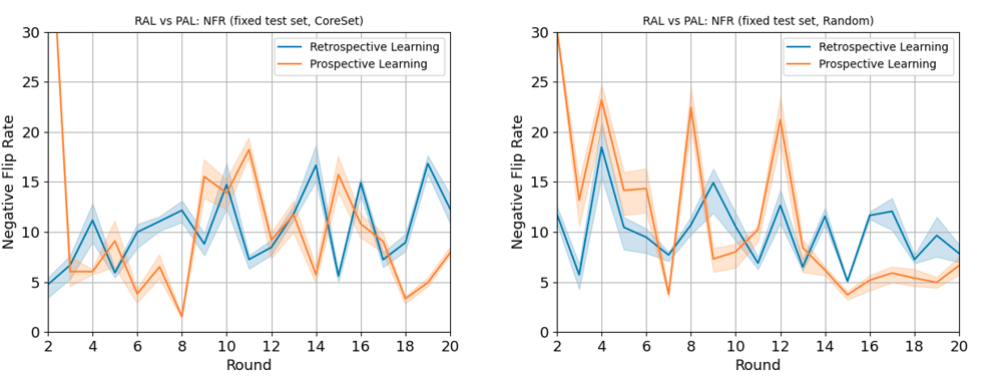}
\caption{NFR for Coreset and Random (High NFR indicates more test samples forgotten)}
\label{fig:nfr}
\end{figure}

Other reasons for regions of lower accuracy when compared to the retrospective setting come from the concept of catastrophic forgetting, which occurs when a model is trained on a set of data samples but loses its ability to perform well on this set when new samples are added. Catastrophic forgetting is characterized by the negative fliprate (NFR) of a model, described as summing the total number of model predictions that changed from correct to incorrect after new data was added to the training set \cite{yan2021positive}. We show NFR results for Random and Coreset in Figure \ref{fig:nfr}.

\begin{figure*}[h]
    \centering
    \includegraphics[width=\textwidth]{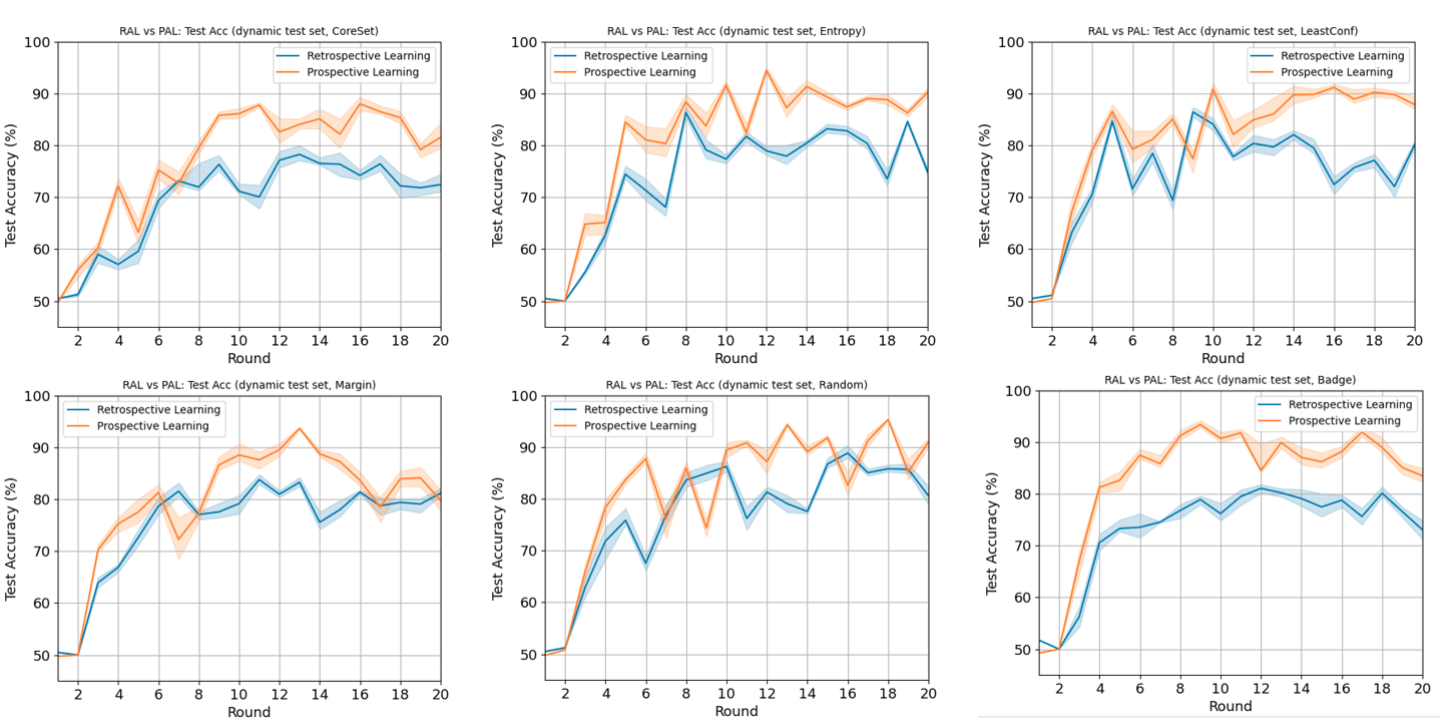}
    \caption{Retrospective active learning (RAL) vs Prospective active learning (PAL) test accuracy results averaged over 5 seeds for dynamic test set. }
    \label{fig:test-acc-dynam}
\end{figure*}
For Figure \ref{fig:nfr}, we commonly observe spikes in the NFR values for the prospective active learning experiments, as well as the retrospective case. In many cases, this appears to correlate with regions of lower accuracy noticed in Figure \ref{fig:test-acc}. For example, when analyzing the Coreset query strategy for the prospective case in Figure \ref{fig:test-acc}, at Round 11 we observe lower accuracy than what was observed in previous rounds, which correlates to Figure \ref{fig:nfr}'s NFR spike for the Coreset query strategy at Round 11. This same correlation is seen in other query strategies, like Random at Round 8 and 12 for the prospective active learning curves. Note that this same trend is present in the retrospective case. For example, Coreset at Round 14 experiences both a region of low test accuracy and a spike in NFR values. Therefore, even though the nature of some of these query strategies may can cause inconsistent regions of accuracy as discussed previously, there appears to be a strong correlation between NFR spikes and test accuracy performance deterioration. It is likely that prospective active learning suffers from areas of high NFR due to the fact that it is being tested on the fixed test set, which contains data from visits the model has not been trained on yet. Therefore, this may cause inconsistency in performance. However, despite some inconsistent regions of accuracy, prospective active learning appears to consistently outperform retrospective active learning after the ninth round of training, demonstrating better generalization capability on the fixed test set.

\subsection{Dynamic Test Set Results}

% here

\subsubsection{Retrospective vs Prospective Active Learning}

We present a comparison between retrospective and prospective active learning for the same six query strategies for the dynamic test set, shown in Figure \ref{fig:test-acc-dynam}. When analyzing the graphs, it becomes apparent that prospective active learning typically outperforms retrospective active learning for most query strategies in terms of test accuracy. This change is even more noticeable when comparing the test accuracy results for the fixed test set in Figure \ref{fig:test-acc}, where the results in Figure \ref{fig:test-acc-dynam} show more consistent regions of high accuracy. This change is most likely due to the fact that the dynamic test set only contains data from visits that the model has been trained on, whereas the fixed test set contains data from visits the model has not learned yet. 

Query strategies like Entropy and BADGE noticeably perform much better than retrospective active learning in Figure \ref{fig:test-acc-dynam}. BADGE, for instance, achieves over $90\%$ accuracy as early as Round 9 using 2,432 training samples. When this same query strategy is applied in the retrospective setting, however, the test accuracy does not surpass $85\%$ throughout all 20 rounds. A similar trend is present in the Entropy plot, where the prospective setting stabilizes to a test accuracy of approximately $90\%$ during later rounds. Meanwhile, the retrospective setting fluctuates frequently in test accuracy during later rounds, appearing to be less stable than the prospective case. Other query strategies like Coreset and Least Confidence exhibit similar performance.

Overall, these results demonstrate that even though retrospective active learning has the entirety of the dataset to select data from, it does not generalize as well to visit-specific data. Prospective active learning, on the other hand, is able to achieve higher accuracy than the retrospective setting by sampling images in a sequential manner, imitating the manner in which these images were collected. These results further validate the claim that the active learning process for clinical trial data must be viewed from a prospective lens.
% to do: badge, least confidence perform well: EXPLAIN WHY; this is also the case for the fixed set

% \subsubsection{Negative Flip Rate}
% We analyze the results presented in Figure \ref{fig:test-acc-dynam} more thoroughly through the NFR, summarized in Figure \ref{fig:nfr-dynam}. Similar to previous results, drops in test accuracy are linked to spikes in NFR. For example, the random query strategy has the most drops in accuracy compared to the other query strategies for the dynamic test set. When looking at Round 7 and Round 9, for instance, we see this correlation. For many of the query strategies, Figure \ref{fig:nfr-dynam} even shows that the NFR curve for prospective active learning is lower than that of retrospective active learning in later rounds. This demonstrates that querying data samples non-sequentially results in more information being lost over time, contributing to the accuracy plots in Figure \ref{fig:test-acc-dynam}.

% Comparing previous NFR results in Figure \ref{fig:nfr} with Figure \ref{fig:nfr-dynam}, the NFR curves for Figure \ref{fig:nfr-dynam} are lower than those in Figure \ref{fig:nfr}, meaning that the model is retaining information previously learned about past visits. 
% \begin{figure*}[h]
%     \centering
%     \includegraphics[width=\textwidth]{Figures/nfr-dynam.png}
%     \caption{Retrospective active learning (RAL) vs Prospective active learning (PAL): NFR results averaged over 5 seeds for dynamic test set. }
%     \label{fig:nfr-dynam}
% \end{figure*}

\section{Conclusion}
In this work, we propose a prospective active learning framework applied to clinical trials where the collected data is known to have interventional and time dependencies. We analyze this method's performance against traditional retrospective active learning. We conclude that by conditioning on the time (or visit, in the case of a clinical trial) of which an image was collected, we enforce the i.i.d. assumption that most active learning query strategies adhere by. Because of this, we observe an increase in performance on two different types of test sets: a fixed test set and dynamic test set. The fixed test set tests overall generalization performance, as it is composed of OCT images from various visits. The dynamic test set expands each training round and examines current and past visit generalization, as it is contains up to the current visit of OCT images the model has been trained on. We use NFR as an additional metric to examine how much information is retained over time, comparing the NFR results to what is observed with the accuracy plots. Furthermore, we perform experiments to understand the impact of the order in which data from visits is sampled. We conclude that when analyzing data collected from a clinical trial, active learning must be performed in a prospective manner, as sampling sequentially is what allows the model to learn the interventions across the training data. In addition, our method mirrors how data is collected in a clinical trial setting, thus introducing more realistic sampling. Future directions for this type of research include further validating this framework on other clinical trial datasets beyond OCT, as well as considering the effect of image alignment within these datasets across patients' visits. Furthermore, incorporating a healthy patient population into the datasets may provide additional insights into the performance of prospective active learning, where our current understanding expects prospective and retrospective active learning to provide similar results if minor changes are observed in the healthy cohort across visits. In addition, the effect of different model architectures and query sizes can be further investigated.

\section{Acknowledgements}
This material is based upon work supported by the National Science Foundation Graduate Research Fellowship under Grant No. DGE-2039655. Any opinion, findings, and conclusions or recommendations expressed in this material are those of the authors(s) and do not necessarily reflect the views of the National Science Foundation.

%%
%% Print the bibliography
%%
\printbibliography

%%
%% If your work has an appendix, this is the place to put it.
\appendix

\end{document}